# Intelligent Acoustic Module for Autonomous Vehicles using Fast Gated Recurrent approach

Raghav Rawat[1]
1. ECE department
RV College of Engineering
Bengaluru, Karnataka, India
raghavrawat.ec18@rvce.edu.in

Shreyash Gupta[1]
1. CSE Department
RV College of Engineering
Bengaluru, Karnataka, India
shreyashgupta.cs18@rvce.edu.in

Shreyas Mohapatra[1]
1. ECE Department
RV College of Engineering
Bengaluru, Karnataka, India
shreyasm.ec18@rvce.edu.in

Sujata Priyambada Mishra[1]
1. Assistant Professor, ECE department
RV College of Engineering
Bengaluru, Karnataka, India
sujatapm@rvce.edu.in

Sreesankar Rajagopal[1]
1.Senior Architect, Engineering R&D
Wipro Limited
Bengaluru, Karnataka, India
sree.sankar@wipro.com

*Abstract—* **This paper elucidates a model for acoustic single and multi-tone classification in resource constrained edge devices. The proposed model is of State-of-the-art Fast Accurate Stable Tiny Gated Recurrent Neural Network. This model has resulted in improved performance metrics and lower size compared to previous hypothesized methods by using lesser parameters with higher efficiency and employment of a noise reduction algorithm. The model is implemented as an acoustic AI module, focused for the application of sound identification, localization, and deployment on AI systems like that of an autonomous car. Further, the inclusion of localization techniques carries the potential of adding a new dimension to the multi-tone classifiers present in autonomous vehicles, as its demand increases in urban cities and developing countries in the future.**

## I. INTRODUCTION

TinyML is machine learning done on low power devices like microcontrollers. These devices do not possess high computing power. Essentially, it is very cheap hardware employing a range of inexpensive smaller components, but together with machine learning capabilities, it is easily the next revolution in IoT and Edge devices. As interest in deep learning research grows, the inclination to use lesser and lesser number of parameters to achieve equivalent performances is becoming the norm, to facilitate these capabilities on the small and smart pocket devices of today. As we witness today's everchanging tech industry TinyML or in general, Edge AI, is indubitably the next go to idea.

GPT-3 from OpenAI in 2020[1] created a revolutionary mark in few shot learning language models, capable of writing everything from a newspaper article, poems, to even blog posts on uncommon topics. Just throw in a request for a thousand words on a random topic and the machine learning will serve you with human-like work in no time. GPT-3 running at 175 billion parameters is a vivid but incredibly complex architecture. These many parameters require their own time to be tweaked and adjusted for the optimum results. This infers the requirement of a lot of time and computing power. But gradually, some researchers have found that these many parameters are not required to achieve a nearly similar performance. In fact, it is proved that just 0.001% of the parameters can solve the purpose. But that still makes for 223 million parameters, which is still a huge number to be able to run on a microcontroller. Thus, research in the recent years have been inclined towards this reduction in ML models to get near state-of-the-art performance, and possibilities of dumping the models on the microcontrollers that always surrounding us and our lives.

RNN Architectures like GRUs [2] and LSTMs [3,4] are well known for a long time now for cutting edge fields like time series prediction, NLP, NLU, text mining and data mining. But larger sequence models do have their own share of limitations. RNNs have shown a trade-off between accuracies and model complexities, leading to slower computation for present day AI challenges. If activation functions are used at each unit, the computation time grows very fast for larger networks. This ultimately leads to exploding and vanishing gradients. Thus, use of long known RNNs implementations becomes baseless for real-time problems of today like text auto-completion, AI chatbots, time series forecasting, machine translation, video tagging, etc. This paper has addressed these issues entailing with RNNs for yet another real-time problem of sound identification. As commonly understood, sound recognition is a use case for small scale devices which are accessible to us throughout a normal workday. Devices like smartphones, laptops and other central processing units like ones built into automobile systems, emergency appliances and security biometrics, when facilitated with sound recognition, can deliver some intriguing and handy applications to ease our daily life. A conclusion from the same can be drawn that sound recognition is a feat that is plausible on edge and IoT devices.

This paper builds up and discusses the application of sound recognition, in perspective of a module for a self-driving car, in three phases. Phase A - establishing need for acoustic sensor networks (ASN). Phase B - Implementation and diagnostics of acoustic deep learning classifier. Phase C - Model Deployment and Test runs for model accuracy on real-time environments.

## II. ACOUSTIC SENSOR NETWORKS

The acoustic sensors are placed at several points inside and outside the vehicle, such positions are determined by calculating vector distances from presumed nodal location to the periphery of the car, which cover the proximal areas of the aforementioned car. With the application of Internet of things (IoT) these myriads of sensors provide a facet of wireless communication which is abbreviated as wireless acoustic sensor network (WASN). These strategically placed sensors have a very specific and targeted objective, they are responsible for collecting and sampling sounds of the streets, other traffic related sounds. The paper describes an ML model where we introduced classes for most frequent sounds on roads like car horns, ambulances, police sirens and other ambient sources that contribute to noise[5], which either needs to be filtered out or eliminated in further stages of the model. Acoustic sensors are generally more adept at dealing with traffic sounds when compared across the same domain as radio frequency devices. As the complexity and hierarchy of an acoustic sensor network increases more favorable and efficient outcomes are produced with relatively high and constant accuracy. They yield high detecting capabilities and provide a whole range of new functions. One ingenious way to achieve such fruitful outcomes is by using two microphone and synchronizing them to detect sounds on the same line of convergence of acoustic waves and also calibrating them such that most common to sounds attributed to noises can be eliminated by intensity levels[6]. Current application spaces like smart urban communities and structures, encompassing helped living, or natural surroundings checking have effectively exhibited the interest for acoustic-based arrangements. Once said sensor networks are placed in the predetermined locations, they help in detecting and notifying the car of several germane proximal developments. These include vehicles around the car by virtue of honking, the direction of arrival of the surrounding car can be established. Further pragmatic applications like the presence of a vehicle at blind turn (no visibility) can be anticipated and notified to the car in dire time. And by applying the rules and calculation according to Doppler's effect more research and features are being developed daily. Time of arrival of proximal moving objects by virtue of the sounds they produce is also a methodology[7]. When composited into a specific model, it provides a new dimension in autonomous cars that further help in creating more advanced autonomous vehicles.

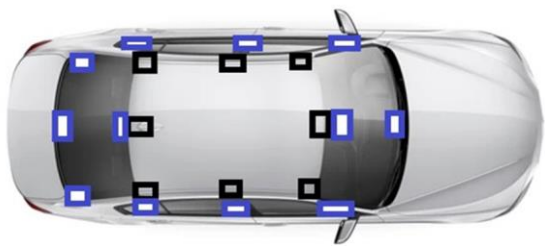

Fig1. Location of acoustic sensors placed symmetrically on the car to minimize noise sounds.

There are several approaches to localization for these sensor networks. The primary criteria include the ability to detect mixed sound samples, not just one single tone and by self-localization the nodes of the network are able to detect their relative vectors, with respect to the source[8]. Other techniques like source localization are also in application with limited success for three-dimension analysis in the case for acoustic models. Hence location of the sensors is quintessential. For this unique predicament, acoustic source localization has to be the most sought-after domain that has attracted interest and subsequent research in topics local to this. In simpler terms, the limitation of acoustic sensors can be accomplished by examining energy and worldly or potentially directional highlights from the approaching sound at various receivers and utilizing an appropriate model (ML as it pertains to this paper) that relates those highlights with the spatial area of the source of interest and also further facilitates by a notification feature to the user[9].

## III. ACOUSTIC LEARNING MODULE

Complex problems like those for a self-driving car need to be addressed by more robust domains of machine learning. The architecture demonstrated through this paper is that of a Recurrent Neural Network. A modification of RNN, Fast Accurate Stable Tiny Gated Recurrent Neural Network (FASTGRNN)[10] has been used on the grounds of its previously proven supremacy for state-of-the-art performance metrics on time series data. Although this paper would also be covering performance comparisons of FASTGRNN with other at-par deep learning architectures and algorithms, but in the end for edge devices and AI on IoT implementations a model which performs equally well even with certain added environmental limitations and resource constraints proves to be the apt one for steady deployment and testing.

Acoustic sensor networks leverage us to capture the real data around and an inter communication, enabling complete surrounding coverage. But a generalized model training and development requires considerations like noise levels, varied locations and varying traffic conditions across continents. In order to prevent the model from overfitting for the recorded data from a specific country, or a locality in the worst case, a new dataset is curated having an amalgamation of independent datasets like UrbanSound8K [11,12] and Emergency-Sounds, and other independent internet sources. A total of 6 different classes of sounds were cultivated with the number of datapoints as given below in table 1. These were then capped at minimum 3 and maximum 4 seconds of duration.

Table1. Datapoints for each class in the curated dataset

| Class Name | Number of samples |
|---|---|
| Car Horn | 932 |
| Children Playing | 939 |
| Dog Bark | 758 |
| Drilling | 790 |
| Engine Idling | 933 |
| Siren | 861 |

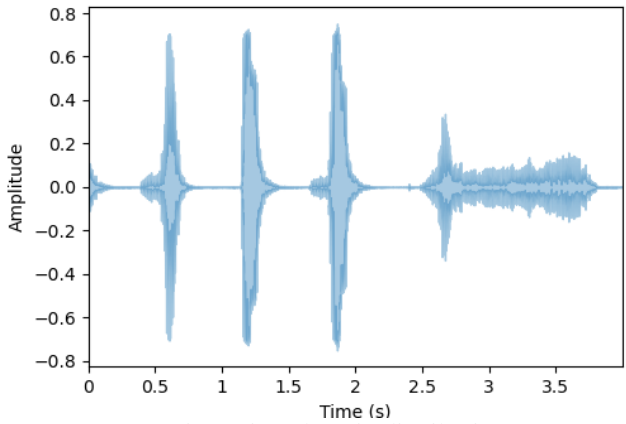

Fig3. Time domain distribution

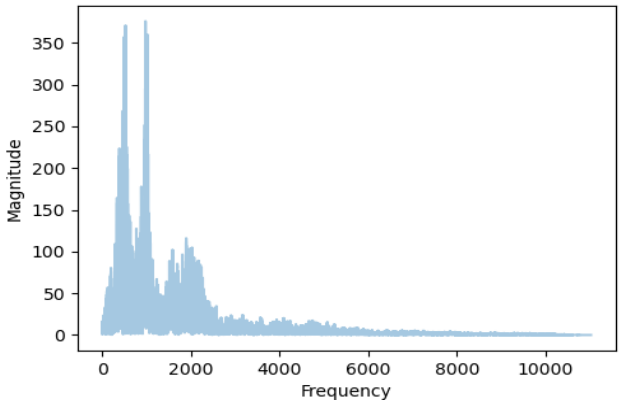

Fig4. Power Spectrum

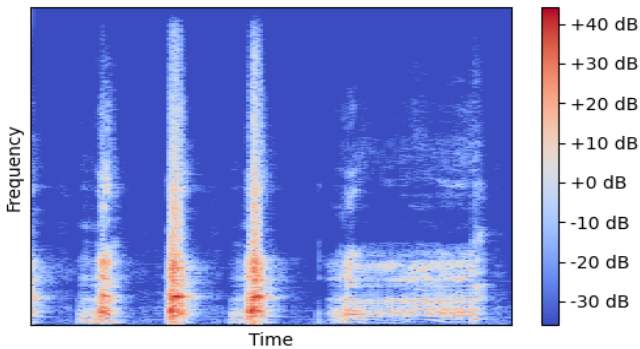

Fig5. Spectrogram Plot

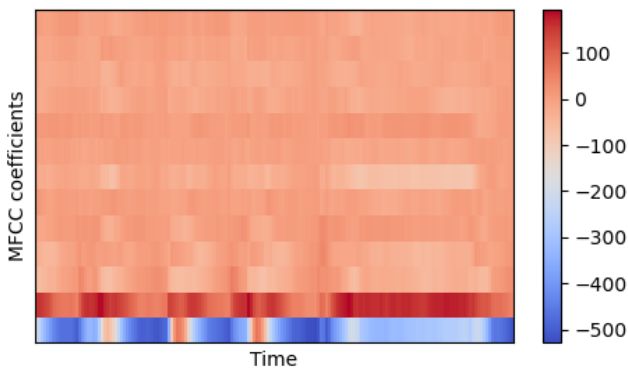

Fig6. MFCC Spectrogram

Fig 3-6 Shows the preprocessing workflow for transformation of time series audio data, of a dog barking, into frequency domain Mel cepstral coefficients (13-dimensional)

The plots from Fig3-6 are constructed using Librosa sound processing library. The sound sample chosen for this representation is that of a Dog bark. Fig3 represents an Amplitude-Time wave plot of the audio signal, sampled at the rate of 22050 samples/second. Following that, Fast Fourier Transform is done on the signal and an absolute value for all complex signals are calculated. Fig4 represents the plot of these absolute values, which is also known as a power spectrum of the signal, which signifies magnitude contributions of each frequency in the range. Since the sample is that of a dog bark, it is expected and concurrently observed from the plot that lower frequency ranges have greater contributions. Next, we perform Short Time Fourier Transforms (STFT) on the signal with a window of fixed samples and hop length for each STFT. Finally, an absolute value of the entire output array of the STFT is taken to plot on a Decibel scale Frequency-Time spectrogram as shown in Fig5. Multiple MFCC coefficients are extracted from segments of fixed duration from the signal, which ultimately gives us our numbered dataset for neural network training.

This base dataset was further used for noise estimation and removal, as well as multi-class prediction. This paper proposes the solution to multi class classification on audio clip with multiple sources by acknowledging presence of a source if final predicted probability for its class exceeds its threshold value. Threshold value for a class can be calculated by averaging out the prediction probability of that class in audio clips from train data where the class is known to occur. Noise removal techniques like spectral gating[13,14,15] were employed on some samples to elucidate the improvements in sound identification with efficient noise filtering pre-processing step.

## IV. DATA PROCESSING

Complex deep learning problems require numerical data as a prior step before feeding it to Artificial Networks as the learning process. In computer vision problems and applications as well, majority work is on the pixel magnitudes and different kernels based on values, which facilitates for the numerical data to infer patterns and ultimately a prediction for the image. On similar lines, in the case of audio data (wav or mp3 files) there is a need to extract numerical patterns of the data well. This issue can be addressed by using signal processing techniques. Audio files are taken as the input for the preprocessing pipeline. This process begins with a fast Fourier transform of the audio files using sound libraries for python which gives us the relative magnitude of importance of all possible frequency components available within a given audio clip. This is known as a Frequency Spectrogram. In the time space, the preparation of signals on computers is more confounded, so it is plausible to rather investigate the highlight features in the frequency domain. Likewise, since the sound information is huge in both time and frequency space, to decrease the measure of information to be handled and processed, the transient highlights and features should be extricated to address the sound segments [16]. This is also known as the windowing technique, where convolutions on smaller segments of audio clips are done to extract statistical frequency domain features. These features in a json format are fed into the Neural Network architectures as depicted in Fig2.

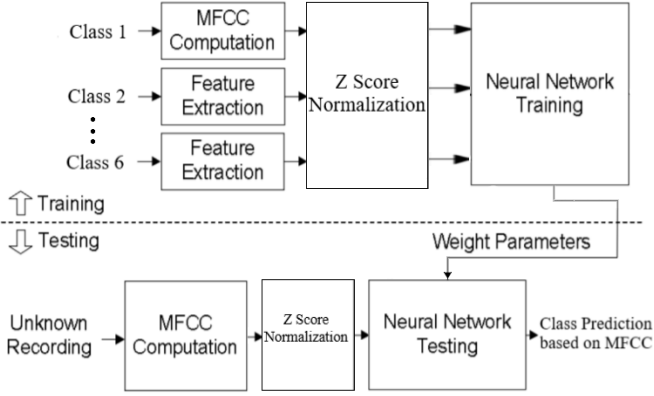

Fig2. Data Processing and Feature Extraction workflow on audio data.

At present, the most generally utilized highlight features in sound signal processing are Perceptual Linear Prediction (PLP) coefficients, Linear Prediction Coefficients (LPC) and Mel-Frequency Cepstral Coefficients (MFCCs)[11]. By and large, MFCCs are utilized as phonological component parameters in speaker identification applications. Further, it has been shown suitable in crisis sound signal identification frameworks effectively[17], and thus MFCCs are utilized in this work as well. For each 3 second clip of a sound signal, 5 segments each and ultimately, on using a standard sampling rate of 22050, 13- dimensional MFCCs are computed for each segment from their spectral representations known as spectrograms. This spectrogram audio example in Fig 3-6 is that of a dog barking, which can also be observed from the loudness distribution which is equally spaced with silent gaps. Starting from Fig3, audio data as time series data is applied a short time Fourier Transform to get magnitude vs Frequency distribution, also known as the power spectrum. The power spectrum(Fig4) signifies the relative presence and contribution of different frequencies and hence a low frequency and high decibel plot for a dog barking. Further Fig5. frequency vs time gives a plot of the audio data in decibel scale. Fig.6. represents the MFCCs extracted into 13-dimensional information bands. Each band can be quantified by a single cepstral coefficient.

## V. MODEL ARCHITECTURE

As an initial approach, dense neural network implementation is done using Keras. 'Relu' activation along with a 30% dropout in hidden layers makes the model prevent overfitting. 'Softmax' activation is used in the final three-way classifier output layer. 'Adam' Optimizer helps in learning the data (MFCC vectors) by tuning weights of the hidden layers. In the case of sequence models LSTMs proved to be of better performance but relatively poor model size and computation cost, resulting them to be impractical for edge devices.

The model architecture consists of 26 Fast GRNN cells connected linearly. Each cell has parameters $U, W, nu, zeta, B_z$ and $B_h$. Except for the first cell, each cell requires output of the previous cell in order to calculate its output. Each cell accepts an input of dimension 13, output of the previous cell of dimension (26,1) and gives out a (26,1) vector using its parameters. FastGRNN architecture uses a scaled residual weighted connections for every individual cell of the hidden state. Final part of the architecture is a fully connected layer consisting of 6 perceptrons. This last layer compiles the data extracted by previous cells to form the final output. 'Softmax' activation function is applied on this last layer to obtain probability of occurrence for each class. FastGRNN[10] intrinsically uses shared parameter matrices U and W to compute the hidden state $h_t$ as well as the gate $z_t$.

$$z_t = \sigma(Wx_t + Uh_{t-1} + b_z) \quad (1)$$
$$\widetilde{h_t} = tanh(Wx_t + Uh_{t-1} + b_h) \quad (2)$$
$$h_t = \left(\zeta(1 - z_t + \nu) \odot \widetilde{h_t} + z_t \odot h_{i-1}\right) \quad (3)$$

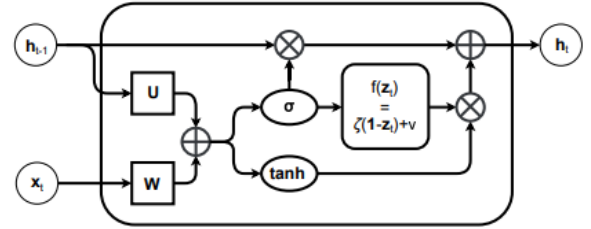

Fig7. GRNN single unit structure [10]

| Complete Learning algorithm in Fast Gated Recurrent Network | | |
|---|---|---|
| Step 1 | : | **Input** Labelled training data .mp3/.wav files, 3 seconds each into 5 segments $--$ (0.6 sec × 5 segments) |
| Step 2 | : | **Fourier Transform** into frequency domain feature extraction $--$ SR = 22050; 13230 Samples/Segment |
| Step 3 | : | **MFCCs** for each segment, (26 = 2 x 13 - dimensional vectors) |
| Step 4 | : | **Data in json** is stored for recurrent network training $--$ ($\vec{x} \in R^n$) where ($x_n \in [+1, -1]$) |
| Step 5 | : | **Parameter Update** $--$ $W, U, b_h$ and $b_z$ |
| | : | $\sigma$ **and** $tanh$ activation |
| | : | **Classical Loss** function computation for optimal loss landscape |
| | : | **Repeat Step 5 until convergence** |
| Step 6 | : | **Updating** GRNN parameters and optimum hyperparameters |
| Step 7 | : | **Segmented inferencing** |

| State-of-the-art Model Architectures Compared | | | |
|---|---|---|---|
|  | **Accuracy % ↑** | **# Of Parameters ↓** | **Size in KB ↓** |
| Standard DNN | 89.96 | 321,734 | 111 |
| CNN with dropout regularization layer | 88.92 | 18,438 | 302 |
| GRU | 89.23 | 278,784 | 1458 |
| LSTM | 89.91 | 330,178 | 1980 |
| FASTGRNN | 87.89 | 1,230 | 4.8 |

In the multi-class RNN structure the target is to learn a function $F : R^{D \times T} \rightarrow \{1, ..., N\}$, that predicts one of the N classes for any given dataset point X. The ordinary RNN structures have the facility to produce an output at each unit and time step, but only the setting where every data point is linked with a single label prediction at the termination of the time constraint T is considered. The ordinary RNNs evaluate and maintain a hidden cell state $h_t \in R^{\hat{D}}$ that captures the temporal sequence characteristics over the input data as shown in equation (2) above. The problem of learning U and W is difficult in ordinary architectures as the gradients can have exponentially increasing numbers. Unitary methods tend to have a control over condition numbers but the requirements of training time can be drastically large or generated model's accuracies can be well below the anticipated range. To minimize the total parameter count, FastGRNN reuses the U and W matrices for even the vectored gated functions of the cells. Therefore, FastGRNN's inferential complexity is the nearly similar to that of standard RNN but its accuracy and stable training features are comparable to expensive GRU and LSTM architectures. It is in turn able to converge in $O(1/\epsilon^2)$ and independent of T. All this results in a prediction cost of GRNN much lower than Unitary and Gated structures.

## VI. PROBABILISTIC INFERENCE

Once training is completed, all parameters of the model $U, W, nu, zeta, B_z$ and $B_h$ are saved in the form of NumPy files. An audio file for format .wav is taken as input. 0.6 seconds of audio containing 22050/5 samples is processed at a time. 338 mfccs are extracted from .6 seconds of that audio clip. Mean and std that were saved at the end of training phase are used to perform Z-Score Normalization on the mfccs. Normalized mfccs are used as inputs of the model's first layer. Outputs of cells are calculated using equations (1,2,3).

$$P = U_{26}W_{FC} + b \quad (4)$$

$$V = \frac{e^{P_i}}{\sum_0^5 e^{P_i}} \quad (5)$$

Outputs of the last cell are passed through a fully connected layer to give 6 outputs as shown by Equation (4). SoftMax function, depicted by Equation (5), is applied on these outputs to get final probabilities of occurrence for each class. The sum of these probabilities is always one. It can be inferred from this resultant array that which class has higher dominance or say the percentage confidence with which the model predicts that class.

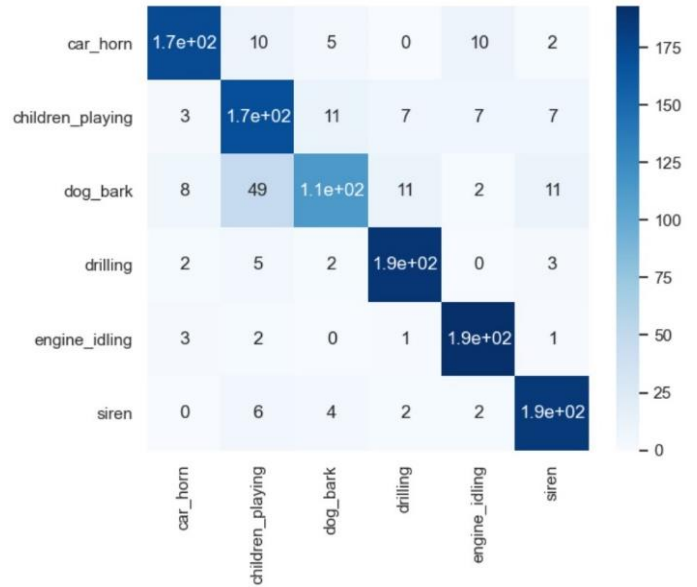

Fig.8. Confusion matrix for trained GRNN model

Fig.8 shows the confusion matrix obtained on running the trained model on test data containing 1200 samples distributed uniformly among the 6 classes. As seen in the vertical scale, greater confidence of the model in predicting class i when the ground truth is j maps to a darker color. The confusion matrix is constructed using Sklearn and Seaborn libraries.

## VII. RESULTS & CONCLUSION

Unitary RNNs balance out RNN training by taking into learning consideration just the very much characterized state change matrices. Thus, there is a limitation in their expressing capability and predictive accuracy even with increasing training time. The training algorithm tends to converge only in small and thus, only a limited accuracy is achieved on most datasets. Accuracy can be somewhat increased by increasing hidden units but again at the expense of training and predicting time needed and model size. Gated architectures also prove the state-of-the-art accuracies by adding extra parameters but also the model size and training time increases manifold.

On the other hand, FastGRNN architecture, achieved by converting the residual connects between cells to a gate while simultaneously reutilizing the RNN parameter matrices, gives a manifold reduction in parameters. A slight decrease in prediction accuracy resulted from FastGRNN's matrices being low-ranked, sporadic and quantized, which allowed the model compression to 2-6 KBs without compensating the accuracy in most cases. At the same time, it also ended up in trained models that can be up to 35 times small in size and have 2-4 times fewer parameters than leading gated RNN techniques like LSTM, GRU, and UGRNN. FastGRNN made predictions 28-42 times faster than other leading RNN models. These characteristics have allowed FastGRNN to be deployed and make predictions effectively on the severely resource-constrained edge and IoT devices like Arduino Uno, ARM Microcontrollers, and Raspberry Pi which were too small to be practically deployed on other RNN models.

## ACKNOWLEDGMENT

We thank Sujata Priyambada Mishra and Sreesankar Rajagopal for providing useful insights and comments on the manuscript as well as overall construction of the work. We thank R.V College of Engineering and Wipro Limited for continued support.